\documentclass[10pt,twocolumn,letterpaper]{article}

\usepackage[pagenumbers]{cvpr} 

\definecolor{cvprblue}{rgb}{0.21,0.49,0.74}

\usepackage[table]{xcolor}
\usepackage{multirow}
\usepackage{pifont}
\newcommand{\cmark}{\ding{51}} 
\newcommand{\xmark}{\ding{55}} 
\usepackage{booktabs}   
\usepackage{multirow}  
\usepackage{makecell}   
\usepackage{siunitx}   
\usepackage{arydshln}
\usepackage{graphicx}
\usepackage{caption}
\usepackage{csquotes}
\usepackage[accsupp]{axessibility}  
\usepackage[pagebackref,breaklinks,colorlinks,allcolors=cvprblue]{hyperref}

\def\confName{CVPR}
\def\confYear{2026}

\title{MedLoc-R1: Performance-Aware Curriculum Reward Scheduling for GRPO-Based Medical Visual Grounding}

\author{
Guangjing Yang$^{1}$\quad
Ziyuan Qin$^{2}$\quad
Chaoran Zhang$^{1}$\quad
Chenlin Du$^{3}$\quad
Jinlin Wang$^{1}$\\
Wanran Sun$^{1}$\quad
Zhenyu Zhang$^{1}$\quad
Bing Ji$^{4}$\quad
Qicheng Lao$^{1\dagger}$\\[2mm]
$^1$Beijing University of Posts and Telecommunications\quad
$^2$Emory University\\
$^3$Peking University\quad
$^4$Shandong University \\
{\small\texttt{\{ygj2018, qicheng.lao\}@bupt.edu.cn} \quad $^\dagger$Corresponding author}
}

\begin{document}
\maketitle
\begin{abstract}
    Medical visual grounding serves as a crucial foundation for fine-grained multimodal reasoning and interpretable clinical decision support.
    Despite recent advances in reinforcement learning (RL) for grounding tasks, existing approaches such as Group Relative Policy Optimization~(GRPO) suffer from severe reward sparsity when directly applied to medical images, primarily due to the inherent difficulty of localizing small or ambiguous regions of interest, which is further exacerbated by the rigid and suboptimal nature of fixed IoU-based reward schemes in RL. This leads to vanishing policy gradients and stagnated optimization, particularly during early training. To address this challenge, we propose \textbf{MedLoc-R1}, a performance-aware reward scheduling framework that progressively tightens the reward criterion in accordance with model readiness. MedLoc-R1 introduces a sliding-window performance tracker and a multi-condition update rule that automatically adjust the reward schedule from dense, easily obtainable signals to stricter, fine-grained localization requirements, while preserving the favorable properties of GRPO without introducing auxiliary networks or additional gradient paths.  
    Experiments on three medical visual grounding benchmarks demonstrate that MedLoc-R1 consistently improves both localization accuracy and training stability over GRPO-based baselines. Our framework offers a general, lightweight, and effective solution for RL-based grounding in high-stakes medical applications. Code \& checkpoints are available at \hyperlink{}{https://github.com/MembrAI/MedLoc-R1}.
\end{abstract}
    
\section{Introduction}
In recent years, reinforcement learning (RL) has demonstrated substantial benefits in training large language models, particularly for aligning model behavior with human preferences and enhancing reasoning capabilities~\citep{openai, rafailov2024directpreferenceoptimizationlanguage,shao2024deepseekmath}. 
Motivated by these successes, several studies have begun exploring RL within medical vision–language settings~\citep{lai2025medr1, xu2025medground}. However, for language-guided visual grounding tasks, an important question remains: do such perception-oriented localization tasks truly benefit from reasoning-centric RL techniques?
Recent works such as Visual-RFT~\citep{liu2025visualrftvisualreinforcementfinetuning}, VLM-R1~\citep{shen2025vlm}, and Med-R1~\citep{lai2025medr1} have shown that GRPO-style RL post-training can indeed yield performance improvements and enhance model generalization. Yet these studies largely emphasize prompt engineering and fixed-threshold IoU-based reward schemes when explaining the perceptual gains introduced by RL. 
We argue that, in medical visual grounding, expecting a large vision–language model (LVLM) to learn accurate and fine-grained lesion localization in a single stage is unrealistic. The difficulty becomes even more pronounced when aligning general-domain LVLMs using post-training RL, as this process resembles asking an individual without medical expertise to localize disease lesions—such a process inherently requires gradual, progressive acquisition of spatial and semantic understanding~\citep{zhou2020unet++,jing2025BoxMed-RL,gao2024boosting,yi2025idpa}.

\begin{figure}[t]
\centering
\includegraphics[width=1.0\linewidth]{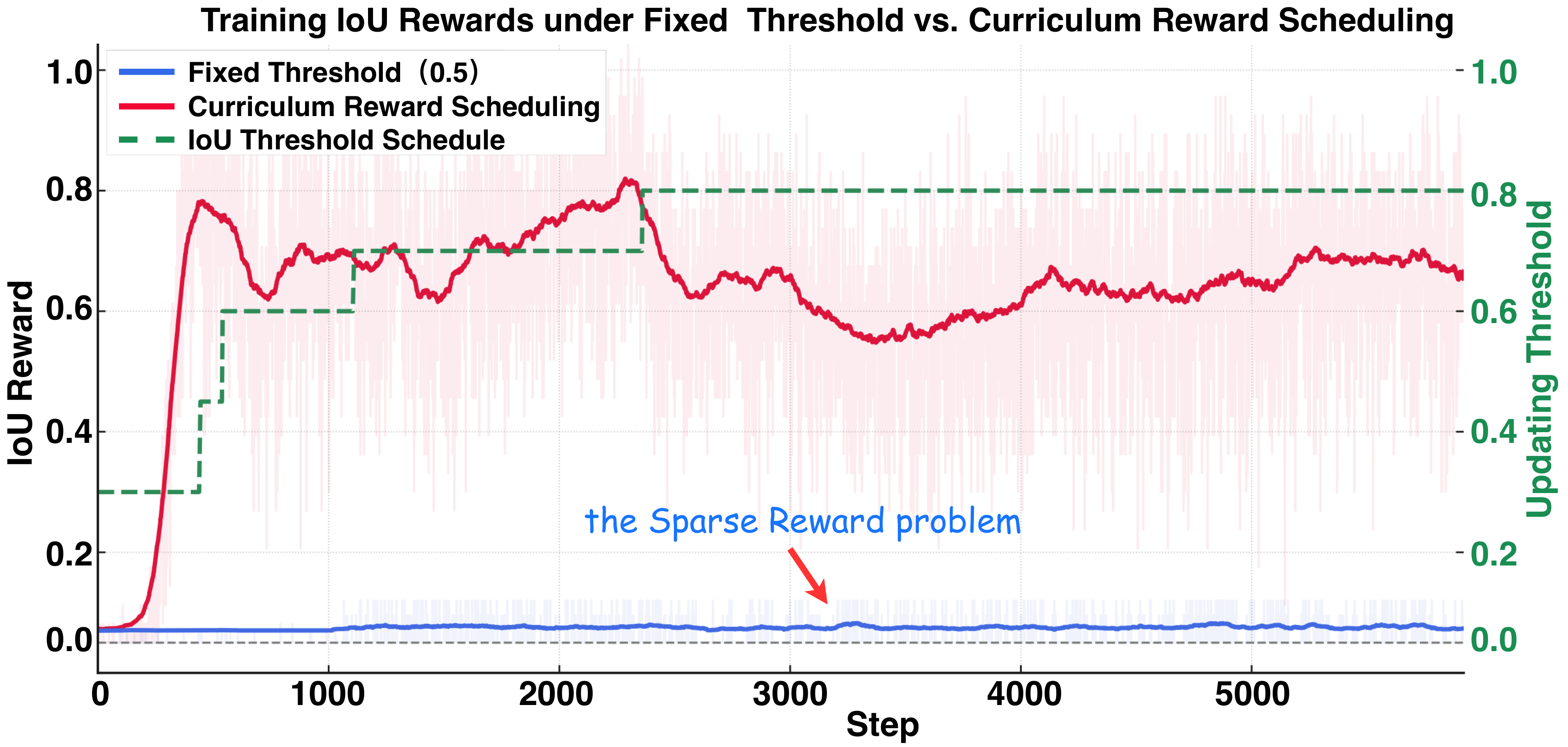}
\vskip -0.1in
\caption{Reward curve with performance-aware progressive reward scheduling (\textit{in red}) showing dense values compared to the reward curve with fixed reward scheme (\textit{in blue}). The dashed line (\textit{in green}) indicates the progression of the reward criterion.}
\label{fig:reward-sparsity}
\vspace{-20pt}
\end{figure}

Unlike natural images, medical images exhibit unique characteristics such as low signal-to-noise ratios, blurred boundaries, small lesion sizes, and severe class imbalance, making the localization problem significantly more challenging~\citep{xu2025medground, rui2025medcco,litjens2017survey,ronneberger2015unet}. 
The recognition difficulty caused by this domain gap often leads the RL policy to experience “frustration,” resulting in training failure to converge, as the model is unable to obtain sufficient positive rewards from multiple attempts. This issue is commonly referred to as the \textit{sparse reward} problem in RL training~\citep{rengarajan2022reinforcementlearningsparserewards}. 
To mitigate reward sparsity and ease the associated training challenges, one intuitive solution is to employ curriculum learning \cite{bengio2009curriculum}, a strategy widely adopted in supervised learning to improve optimization by scheduling task difficulty. However, conventional curriculum methods typically rely on sample reordering or progressive data exposure~\citep{narvekar2020curriculum}—mechanisms that do not directly apply to RL-based grounding, where task difficulty is governed by reward design rather than the input distribution~\citep{ivanovic2018barc}. Moreover, prior RL approaches for visual grounding typically employ a fixed IoU-based reward scheme that cannot automatically adapt its difficulty to the model’s evolving readiness during training ~\citep{shen2025vlm,xu2025medground}. 
As a result, how to construct a progressive and performance-aware reward curriculum remains an open problem.

Motivated by the core principle of curriculum learning—gradual and structured progression—and recognizing the central importance of reward shaping in medical visual grounding, we propose \textbf{MedLoc-R1}, a \textit{performance-aware curriculum reward scheduling} framework that dynamically adjusts the reward criterion according to policy performance. Our scheduling strategy gradually raises the reward criterion as training progresses, enabling a transition from dense to stable rewards, thereby resembling the idea of curriculum learning. Figure~\ref{fig:reward-sparsity} shows this pattern: the \textit{red} reward curve shows progressive reward accumulation along with the curriculum reward schedule level in the \textit{green} dashed line. In contrast, the \textit{blue} curve without reward scheduling suffers from persistently sparse rewards.

Specifically, our method introduces a sliding-window performance tracking module to quantify recent training dynamics. Based on state tracking statistics, we define a multi-condition update criterion that progressively elevates the strictness of the reward criterion, enabling a smooth transition from ``dense rewards---coarse localization'' to ``sparse rewards---fine-grained alignment.'' This design preserves the desirable properties of GRPO while effectively addressing the reward sparsity bottleneck—achieved without introducing auxiliary networks or gradient paths.

Overall, our contribution can be concluded as follows:

\begin{itemize}
\item We identify and formally analyze the reward sparsity problem inherent in applying GRPO to medical visual grounding, clearly revealing how traditional fixed-criterion reward designs lead to unstable optimization and severely degraded policy gradient estimation. 
\item We propose a novel curriculum reward scheduling framework that leverages sliding-window performance statistics to dynamically adjust reward strictness in accordance with model readiness, enabling progressive difficulty control and effectively mitigating vanishing gradients.
\item We conduct extensive experiments across multiple medical grounding benchmarks, demonstrating that our approach significantly improves both training reward dynamics and localization accuracy, while introducing negligible computational overhead.
\end{itemize}
\vspace{-2.0cm}
\section{Related Work}
\paragraph{Reinforcement Learning and GRPO in Medical Visual Grounding.}
Medical Visual Grounding task is a set-free~\citep{clip,glip,gdino,zheng2024curriculum} cross-modal task that locates a medical entity with a given text prompt. Previous work~\citep{qin2023medicalimageunderstandingpretrained, structglip} have shown proper prompts can improve the location performance of LVLMs.
Recent advances in reinforcement learning, particularly value-free methods such as Group Relative Policy Optimization (GRPO) \cite{shao2024deepseekmath}, have demonstrated strong optimization stability in visual grounding and multimodal tasks on natural images. This genre work~\citep{yu2025dapoopensourcellmreinforcement, sun2025tinyr132bpreviewboostingaccuracybranchmerge} proposed to use a rule-based reward to give clear and robust reward signal to replace the value estimation function in PPO~\citep{schulman2017ppo}.
Motivated by these successes, several studies~\citep{lai2025medr1, xu2025medground,yang2026improving} have explored GRPO-based training in medical image scenarios. For example, MedGround‑R1 \cite{xu2025medground} introduces spatial-semantic rewards, while Med‑R1 \cite{lai2025medr1} enhances multimodal diagnostic reasoning. However, these approaches uniformly adopt fixed IoU thresholds for reward design and fail to address the severe reward sparsity present in early stages of medical grounding, which often results in vanishing gradients and slow policy convergence.

\textbf{Curriculum Learning and Reward Sparsity in RL-based Localization.} 
Reward sparsity~\citep{rengarajan2022reinforcementlearningsparserewards,ng2000shaping} is a well-known issue in training RL models, leading to failure. Reward shaping~\citep{andrychowicz2018hindsightexperiencereplay,jaderberg2019human} is a common mitigation strategy for reward sparsity. Curriculum learning~\citep{bengio2009curriculum} is a training strategy that gradually exposes a model to tasks of increasing difficulty, and can used for alleviate reward sparsity~\citep{saito-2018-curriculum,florensa2017reverse}.
Curriculum learning (CL) has been applied to medical imaging tasks for staged sample organization \cite{liu2021style,yuan2025vl,du2024prompting}, and recent work such as MedCCO \cite{rui2025medcco} incorporates CL-inspired strategies to improve multimodal reasoning in RL-based frameworks. Nevertheless, these methods primarily rely on data ordering or progressive sample exposure, which are not directly applicable to localization tasks where difficulty is governed by reward structure~\citep{narvekar2020curriculum}. Existing RL curriculum scheduling approaches \cite{ma2025vtriune,yang2026improving} have not fully addressed adaptive IoU threshold adjustment, leaving reward sparsity and convergence stagnation largely unresolved in RL-driven medical visual grounding.

\begin{figure*}[ht]
	\begin{center}
		\includegraphics[width=1\linewidth]{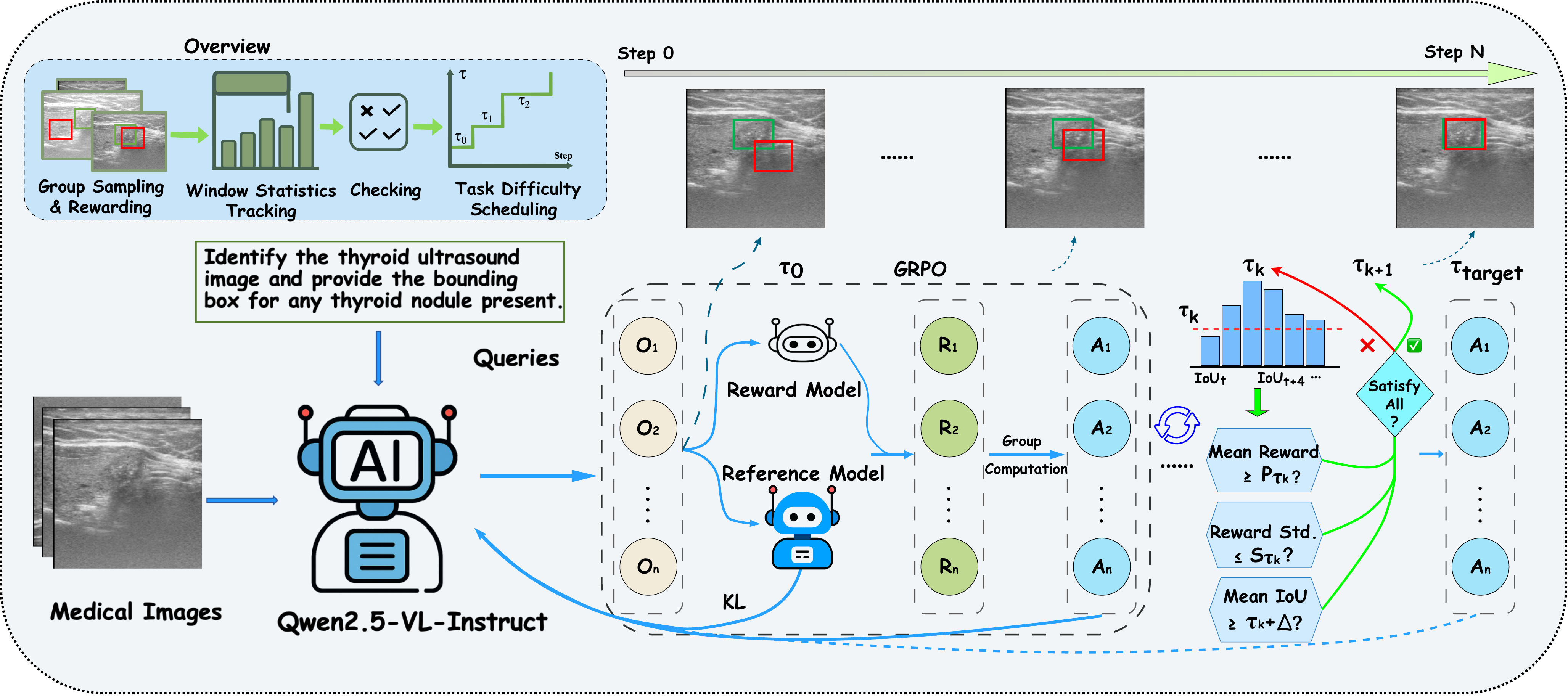}
	\end{center}
  	  \vspace{-0.4cm}
	\caption{Overview of our proposed MedLoc-R1. We propose a progressive curriculum reward scheduling strategy, driven by tracking performance statistics, including mean reward $\bar{r}_k$, reward std. $\sigma_{r,k}$, and the mean IoU $\bar{m}_k$ assessing localization quality.  }
	\label{fig:main_figure}
  \vspace{-0.6cm}
\end{figure*}

\section{Methodology}
\subsection{Preliminaries}
\subsubsection{Group Relative Policy Optimization (GRPO)}

GRPO is a recent variant of Proximal Policy Optimization (PPO)~\citep{schulman2017ppo} designed to remove explicit value function estimation. Unlike traditional PPO, which relies on a critic network, GRPO computes policy gradients by exploiting relative reward differences within groups of actions, thereby reducing variance without additional value networks.
Given an input $x$, GRPO samples a group of $G$ candidate actions $\{a_i\}_{i=1}^G$ from the old policy, each associated with an external reward $r_i$. The normalized advantage $A_i$ for action $a_i$ is defined as:
\begin{align}
A_i &= \frac{r_i - mean(\mathbf{r})}{std(\mathbf{r})} = \frac{r_i - \frac{1}{G}\sum_{j=1}^{G}r_j}{\sqrt{\frac{1}{G}\sum_{j=1}^{G}(r_j-\bar{r})^2+\gamma}}, 
\end{align}
where $\mathbf{r} = (r_1, r2,..,r_G)$ represents the group of rewards for each action in the group $\mathbf{G}$, and $\bar{r}$ denotes the group mean reward and $\gamma$ is a stability term. 
Also, $r_i$ is the final reward received by candidate action $a_i$. 
Then, its optimization objective adopts the clipped PPO form with KL regularization:
\begin{equation}
\begin{aligned}
\mathcal{J}_{\mathrm{GRPO}}(\theta) =
&\frac{1}{G}\sum_{i=1}^G
\min\Big[\rho_i A_i,\ 
\text{clip}(\rho_i,1-\epsilon,1+\epsilon)A_i\Big] \\
&- \beta\, \text{KL}\big[\pi_\theta \,\|\, \pi_{\text{ref}}\big],
\end{aligned}
\end{equation}
where $\rho_i = \frac{\pi_\theta(a_i|s)}{\pi_{\theta_{\text{old}}}(a_i|s)}$, $\epsilon$ is a small constant value for clipping, $\pi_{\text{ref}}$ is a pre-trained reference policy for KL regularization, and $\beta$ controls the regularization strength.

\subsubsection{Problem Definition: Reward Sparsity}

For the visual grounding task, a common reward function is based on whether the predicted bounding box exceeds a fixed IoU threshold $\tau$ with the ground-truth box $b^*$:
\begin{equation}
r_i = \mathbb{I}[\mathrm{IoU}(b_i, b^*) \ge \tau],
\end{equation}
where $\mathbb{I}[\cdot]$ is the indicator function. This reward function will only reward the models when the predicted coordinates contour an area that has a high IoU with the target area. While effective on natural images, this binary reward becomes problematic in medical image analysis, where small lesions, low contrast, and blurred boundaries make it difficult to meet fixed thresholds (e.g., $\tau=0.5$). As a result, for a given threshold $\tau$, we say that all $r_i \in \mathbf{r}$ equals 0. Formally, we have the following definition of this reward sparsity problem in visual grounding:
\begin{equation}
\label{eq:sparsity reward}
\exists \, \tau \in \mathbb{R}, \; \text{s.t.} \forall i \in G,\; 
r_i = \mathbb{I}[\mathrm{IoU}(b_i, b^*) \ge \tau] = 0.
\end{equation}
Hence, the intra-group mean and variance goes to zero:
\begin{equation}
\bar{r} = 0,\ \sigma_r = 0\ \Rightarrow A_i = \frac{r_i - mean(\mathbf{r})}{std(\mathbf{r})}= 0,
\end{equation}
which leads to vanishing policy gradients and training stagnation under GRPO, thereby motivating the adaptive curriculum strategy proposed in this work.

\subsection{Performance-Aware Curriculum Reward Scheduling}
\subsubsection{Task Formulation and Method Overview}

The problem we investigate can be formulated as follows: given an image $I$ and a corresponding query instruction $q$, our goal is to train a vision-language model (VLM) using GRPO to generate reasoning explanations and predict a bounding box $\hat{b} \in \mathbb{R}^4$ for accurate localization of the target region. 

Formally, let the dataset be $\mathcal{D} = \{(I_i, q_i, b_i^*)\}_{i=1}^N$, where $I_i$ denotes the $i$-th image, $q_i$ is the associated query in text form, and $b_i^* = [x^*_1, y^*_1, x^*_2, y^*_2]$ represents the ground truth bounding box. Our objective is to learn a parameterized model $\pi_\theta$ that maps $(I, q)$ to both a bounding box prediction and a coherent reasoning process.

As discussed in the last section, the fixed threshold setting would lead to a reward sparsity issue. Therefore, we propose an approach that employs an adaptive and dynamic reward scheme.
Figure~\ref{fig:main_figure} gives an overview of MedLoc-R1. Our approach comprises two key components: (1) \textbf{sliding-window performance and state tracking}, which monitors recent training dynamics to provide reliable adaptation signals; and (2) \textbf{progressive task difficulty regulation and curriculum scheduling}, which leverages these signals to adjust task difficulty in a gradual and stable manner. Together, these components ensure stable gradient feedback throughout the learning process while continuously driving the model toward higher precision.

\subsubsection{Sliding-Window Performance and State Tracking}

Accurately assessing whether the model has adequately adapted to the current difficulty level is critical for enabling adaptive curriculum scheduling. To achieve this, we introduce a sliding-window-based performance tracking mechanism that continuously monitors recent training dynamics and record the statistics. This mechanism serves two main purposes: (1) quantitatively evaluate the policy’s current capability to meet the IoU threshold, and (2) ensure reliability and stability before increasing task difficulty.

Specifically, at each training step $k$, we maintain a sliding time window $W_k=\{k-N+1,\dots,k\}$ of size $N$ to collect prediction results and corresponding rewards generated by the policy in the most recent $N$ training steps. For step $t\in W_k$, the model generates $G$ predicted boxes $\{\hat{b}_i^{(t)}\}_{i=1}^G$ based on input images, with IoUs between these boxes and the corresponding ground-truth box $b^{*(t)}$ being $\mathrm{IoU}(\hat{b}_i^{(t)},b^{*(t)})$. Given the current IoU threshold $\tau_k$, we define the following three core metrics: 
\par \textbf{(a) Window Mean Reward ($\bar{r}_k$). }  
This metric quantifies the proportion of successful predictions within the recent sliding window, serving as an indicator of the policy's overall performance under the current threshold. Because the reward function is binary (IoU exceeds $\tau_k$ or not), $\bar{r}_k$ effectively estimates the policy’s hit rate:
\begin{equation}
\bar{r}_k = \frac{1}{N}\sum_{t\in W_k} \frac{1}{G} \sum_{i=1}^G \mathbb{I} \Big[ \mathrm{IoU}(\hat{b}_i^{(t)}, b^{*(t)}) \ge \tau_k \Big].
\end{equation}
A higher $\bar{r}_k$ indicates that the current difficulty level no longer imposes substantial challenges, suggesting readiness for progression to a stricter threshold.

\par \textbf{(b) Reward Standard Deviation ($\sigma_{r,k}$). }  
While average reward provides a general measure of success, it is insufficient to assess the consistency of the model’s behavior. To prevent premature threshold escalation caused by transient reward spikes, we incorporate the Reward Standard Deviation metric:
\begin{equation}
\sigma_{r,k} = \sqrt{ \frac{1}{N} \sum_{t\in W_k} \Big( \frac{1}{G} \sum_{i=1}^G r_i^{(t)} - \bar{r}_k \Big)^2 }.
\end{equation}
This metric measures the consistency of the policy’s outputs over recent steps. A large $\sigma_{r,k}$ indicates unstable performance, with success on some samples and failure on others, and thus delays threshold updates. Incorporating this criterion ensures that progression occurs only when performance is both accurate and consistent, promoting stability in curriculum scheduling.

\par \textbf{(c) IoU Margin ($\bar{m}_k$). }  
To ensure that the model exceeds, rather than merely meets, the current threshold $\tau_k$, we introduce the IoU Margin metric, which measures the average surplus over the current IoU threshold:
\begin{equation}
\bar{m}_k = \frac{1}{N} \sum_{t\in W_k} \frac{1}{G} \sum_{i=1}^G  \mathrm{IoU}(\hat{b}_i^{(t)}, b^{*(t)}) - \tau_k. 
\end{equation}
This metric emphasizes whether the policy has the potential to surpass the current stage. If $\bar{m}_k$ is significantly greater than 0, it indicates that the policy's average output IoU is already much higher than threshold $\tau_k$, possessing the capability to accept higher difficulty training.

Compared to binary reward signals, this metric provides a finer-grained assessment of model capability and mitigates stagnation, where the policy barely meets current objectives without achieving substantive progress.

These three metrics jointly constitute the evaluation basis of our reward scheduling mechanism, providing interpretable and controllable signals for curriculum scheduling.

\subsubsection{Progressive Difficulty Regulation and Curriculum Scheduling}
Building on the performance tracking signals introduced in the previous section, we now address the challenge of \textbf{\emph{when and how}} to increase task difficulty. Fixed IoU thresholds are inherently misaligned with the dynamic nature of policy learning: early-stage thresholds that are too strict result in reward sparsity issue, while overly lenient thresholds in later stages fail to enforce fine-grained localization. To overcome this, we adopt an adaptive mechanism that schedules difficulty progression based on model readiness rather than a predefined schedule. Specifically, we define a composite update criterion that integrates the three previously introduced indicators:
\begin{equation}
\mathcal{C}_k := (\bar{r}_k \ge P_{\tau_k}) \wedge (\sigma_{r,k} \le S_{\tau_k}) \wedge (\bar{m}_k \ge \Delta),
\end{equation}
where $P_{\tau_k}$, $S_{\tau_k}$, and $\Delta$ represent the minimum acceptable average reward, the maximum allowable reward variance, and the lower bound on IoU margin, respectively. This joint condition ensures that threshold updates occur only when the policy demonstrates accuracy, stability, and surplus capability, preventing premature difficulty escalation.

To maintain a progressive yet stable learning schedule, these thresholds are dynamically adapted across stages: $P_{\tau_k}$ is gradually relaxed as $\tau_k$ increases, allowing the policy to continue improving even under stricter thresholds; $S_{\tau_k}$ is stage-wise increased to tolerate higher variance under stricter conditions; and $\Delta$ is held constant (e.g., 0.10) for stable margin requirements throughout the training process.

Once the update condition \(\mathcal{C}_k\) is satisfied, the system considers the current stage training to have 
converged, triggering IoU threshold updates using:
\begin{equation}
\tau_{k+1} = \min\bigl(\tau_k + \delta(\tau_k),\; \tau_{\text{target}}\bigr),
\end{equation}
where \(\delta(\tau_k)\) controls the magnitude of threshold increments.
By default, we adopt the following \textbf{\textit{Piecewise Decay}}:
\begin{equation}
\begin{aligned}
\delta_{\text{piecewise}}(\tau_k) =\; & \delta^{(1)}
- (\delta^{(1)}{-}\delta^{(2)})\,\mathbb{I}[\tau_k \ge \beta^{(1)}] \\
& - (\delta^{(2)}{-}\delta^{(3)})\,\mathbb{I}[\tau_k \ge \beta^{(2)}],
\end{aligned}
\end{equation}
with \(\delta^{(1)}\!\ge\!\delta^{(2)}\!\ge\!\delta^{(3)}\!>\!0\) denoting stage-wise step sizes
and \(\beta^{(1)}, \beta^{(2)}\) the boundary thresholds. Beyond this piecewise schedule, we also consider two continuous alternatives that 
require only a single hyperparameter \(\delta_0\) on top of the necessary \(\tau_0\) and \(\tau_{\text{target}}\).
A \textbf{\textit{Linear Decay}} strategy is defined as:
\begin{equation}
\delta_{\text{linear}}(\tau_k) = \delta_0 \cdot \left(1 - \frac{\tau_k - \tau_0}{\tau_{\text{target}} - \tau_0}\right),
\end{equation}
while a \textbf{\textit{Cosine Decay}} variant provides a smoother transition:
\begin{equation}
\delta_{\text{cosine}}(\tau_k) = \frac{\delta_0}{2} \cdot 
\left(1 + \cos\left(\pi \cdot \frac{\tau_k - \tau_0}{\tau_{\text{target}} - \tau_0}\right)\right).
\end{equation}

All three forms preserve fast progression at early stages and fine-grained refinement near the target threshold, ensuring both efficiency and stability during curriculum scheduling. If $\mathcal{C}_k$ is not met, the threshold remains unchanged and the policy continues to optimize under the current difficulty until the criterion is satisfied again.

To prevent outdated training distributions from biasing subsequent evaluations, we partially refresh the sliding window upon each threshold update. Specifically, the earliest half of samples in $W_k$ are discarded, retaining only the most recent $N/2$ steps, and the remaining half is filled with newly collected data to form $W_{k+1}$. 
This strategy enables rapid adaptation to the reward distribution under the new threshold while preserving sufficient historical information to avoid over-sensitivity to transient fluctuations. 
Compared to a full or quarter replacement, this half-retention mechanism demonstrates superior stability in continuous multi-stage scheduling.
Notably, this refresh operation only affects the computation of statistical metrics and leaves the policy update path untouched, ensuring training continuity. Our method integrates curriculum learning into policy optimization to alleviate reward sparsity and improve localization accuracy, all while introducing no extra parameters or computational overhead.

\section{Experiments}

\begin{table*}[ht!]
  \centering
  \footnotesize 
  \setlength{\tabcolsep}{3pt} 
  \caption{Performance comparison across datasets and methods. A@0.5 and A@0.8 denote accuracy at IoU thresholds 0.5 and 0.8. All results are reported over 3 independent runs with different random seeds (42, 43 and 44). \textbf{Bold} indicates the best performance and \underline{underline} indicates the second-best performance among trained methods. Statistical significance: $^{*}$ p $<$ 0.05, $^{**}$ p $<$ 0.01, $^{***}$ p $<$ 0.001.}
  \vspace{-5pt}
  \begin{tabular}{l*{3}{ccc}}
    \toprule
    \multirow{2}{*}{Methods}
      & \multicolumn{3}{c}{\textbf{HAM10000}}
      & \multicolumn{3}{c}{\textbf{HEEL}}
      & \multicolumn{3}{c}{\textbf{TN3K}} \\
    \cmidrule(lr){2-4} \cmidrule(lr){5-7} \cmidrule(lr){8-10}
     &A@0.5  & A@0.8  & mAP
     & A@0.5  & A@0.8  & mAP
     & A@0.5  & A@0.8  & mAP \\
    \midrule

    \addlinespace[1pt]
    \quad Zero Shot-3B   &35.75  &3.39  &12.38  &13.64  &1.64  &4.78  &9.74  &1.29  &3.46 \\
    \quad Zero Shot-7B   &51.42  &11.83 &24.31  &41.79  &2.57  &14.79 &19.12 &2.02 &7.55 \\
    \quad Zero Shot-32B  &66.75  &17.92 &31.01  &46.39  &4.37  &16.92 &21.42 &3.01 &8.71 \\
    
    \addlinespace[3pt]
    \midrule
    \addlinespace[1pt]
    \quad SFT-3B           &90.31$\pm$1.24  &74.22$\pm$2.15  &71.08$\pm$1.87  &92.01$\pm$1.43  &45.18$\pm$3.42  &56.25$\pm$2.93  &62.39$\pm$2.78  &\underline{28.71}$\pm$2.84  &36.11$\pm$2.45 \\
    \quad VLM-R1-3B        &64.65$\pm$2.93  &18.57$\pm$1.98  &31.89$\pm$2.45  &66.79$\pm$3.21  &4.17$\pm$0.73   &21.41$\pm$2.16  &43.01$\pm$2.87  &10.85$\pm$1.54  &20.78$\pm$1.93 \\
    \quad $V$-Triune-3B    &88.92$\pm$1.67  &64.35$\pm$2.84  &65.48$\pm$2.12  &67.63$\pm$3.45  &25.05$\pm$2.73  &38.61$\pm$2.87  &43.50$\pm$2.65  &15.62$\pm$1.87  &21.85$\pm$2.03 \\
    \quad Raw-IoU-3B    &92.86$\pm$3.17  &69.25$\pm$3.15  &66.71$\pm$3.01  &74.24$\pm$3.71  &11.49$\pm$3.19  &35.29$\pm$3.07  &57.17$\pm$2.31  &21.88$\pm$2.57  &29.28$\pm$2.41 \\
    
    \addlinespace[2pt]
    \midrule
    \addlinespace[1pt]
    \quad MedLoc-R1-3B\textsubscript{piecewise}     &94.46$\pm$0.89  &76.02$\pm$1.43  &73.91$\pm$1.12  &94.19$\pm$1.94  &47.35$\pm$2.67  &59.01$\pm$1.89  &66.18$\pm$1.94  &\textbf{29.60}$\pm$2.15  &37.96$\pm$1.67 \\

    \quad MedLoc-R1-7B\textsubscript{piecewise}     &\underline{96.70}$\pm$0.67 &\underline{78.98}$\pm$1.23 &\underline{76.21}$\pm$0.94  &\textbf{96.34}$\pm$0.78  &\underline{57.83}$\pm$2.34  &\underline{64.61}$\pm$1.45  &\underline{67.11}$\pm$2.76  &27.02$\pm$1.98  &\underline{38.21}$\pm$1.43 \\
    \quad MedLoc-R1-32B\textsubscript{piecewise}    &\textbf{97.20}$\pm$3.21 &\textbf{79.01}$\pm$3.23 &\textbf{76.61}$\pm$1.54  &\underline{96.27}$\pm$2.71  &\textbf{57.96}$\pm$3.14  &\textbf{64.80}$\pm$2.15  &\textbf{68.01}$\pm$3.63  &27.52$\pm$3.87  &\textbf{38.71}$\pm$3.79 \\
    \hdashline
    \addlinespace[2pt]
    \quad MedLoc-R1-3B\textsubscript{linear}    &92.51$\pm$2.19 &55.67$\pm$2.36 &60.86$\pm$2.97  
    &92.05$\pm$2.69  &42.17$\pm$3.31  &54.90$\pm$2.15  &62.15$\pm$3.11  &26.97$\pm$3.19  &36.67$\pm$3.99 \\
    \quad MedLoc-R1-3B\textsubscript{cosine}    &93.96$\pm$3.73 &72.79$\pm$3.98 &70.71$\pm$2.45  &93.31$\pm$2.10  &47.67$\pm$3.61  &57.42$\pm$3.83  &65.89$\pm$3.19  &27.21$\pm$2.71  &37.01$\pm$3.64 \\
    \addlinespace[2pt]
    \midrule
    \addlinespace[1pt]
    \quad$\Delta$ (MedLoc-VLM)      &\cellcolor{gray!15} +29.81$^{*}$  &\cellcolor{gray!15} +57.45$^{**}$  &\cellcolor{gray!15} +42.02$^{***}$  &\cellcolor{gray!15} +27.40$^{***}$  &\cellcolor{gray!15} +43.18$^{**}$  &\cellcolor{gray!15} +37.60$^{***}$  &\cellcolor{gray!15} +23.17$^{***}$  &\cellcolor{gray!15} +18.75$^{**}$  &\cellcolor{gray!15} +17.18$^{**}$ \\
    \quad$\Delta$ (MedLoc-VTriune) &\cellcolor{gray!15} +5.54$^{**}$    &\cellcolor{gray!15} +11.67$^{**}$   &\cellcolor{gray!15} +8.43$^{**}$    &\cellcolor{gray!15} +26.56$^{***}$  &\cellcolor{gray!15} +22.30$^{**}$  &\cellcolor{gray!15} +20.40$^{***}$  &\cellcolor{gray!15} +22.68$^{*}$  &\cellcolor{gray!15} +13.98$^{**}$  &\cellcolor{gray!15} +16.11$^{**}$ \\
    \quad$\Delta$ (MedLoc-RawIoU)  &\cellcolor{gray!15} +1.60$^{**}$    &\cellcolor{gray!15} +6.77$^{**}$   &\cellcolor{gray!15} +7.20$^{**}$    &\cellcolor{gray!15} +19.95$^{**}$  &\cellcolor{gray!15} +35.86$^{***}$  &\cellcolor{gray!15} +23.72$^{**}$  &\cellcolor{gray!15} +9.01$^{**}$  &\cellcolor{gray!15} +6.72$^{**}$  &\cellcolor{gray!15} +8.68$^{***}$ \\
   
    \bottomrule
  \end{tabular}
  \vspace{-3pt}
  \label{tab:main_table} \vspace{-5pt}
\end{table*}

\begin{figure*}[t]
    \centering
    \includegraphics[width=1.0\linewidth]{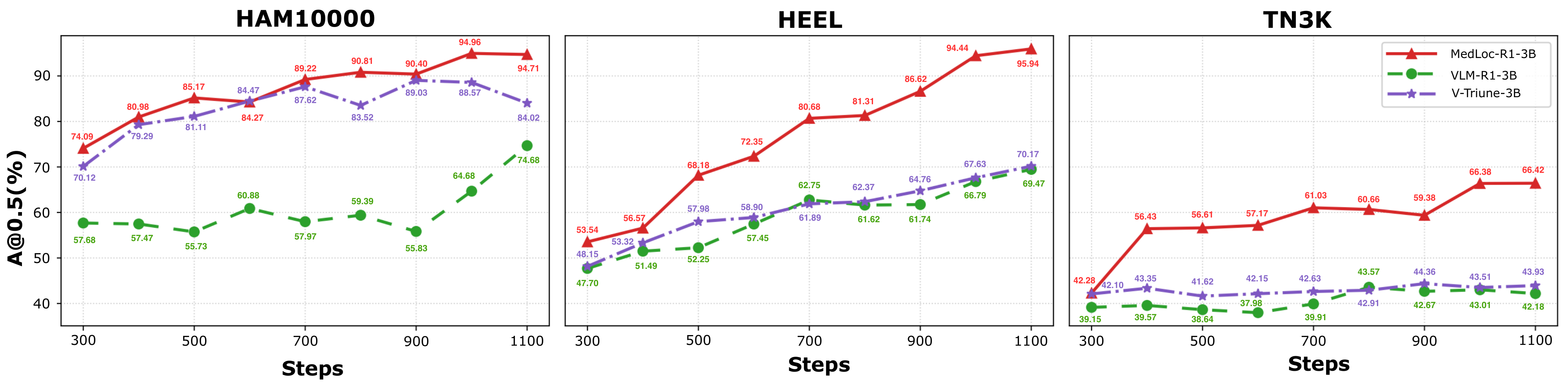}
    \vspace{-20pt}
    \caption{\textbf{A@0.5 (\%) performance across adjacent training steps} on HAM10000, HEEL, and TN3K. Each subplot compares the proposed MedLoc-R1-3B model with two baselines. MedLoc-R1-3B consistently achieves higher A@0.5 and exhibits stronger gains with increasing steps, while V-Triune-3B shows moderate improvement and VLM-R1-3B remains the weakest baseline.}
    \label{fig:performance_over_stpes}
    \vspace{-15pt}
\end{figure*}

\subsection{Experimental Setup}

\textbf{Datasets.} We evaluate our proposal on three medical image grounding datasets of diverse imaging modalities: HAM10000 (dermoscopy), HEEL (X‑ray), and TN3K (ultrasound)~\cite{tschandl2018ham10000, ali2024medcapsnet, gong2021multi-task}. As these datasets were not initially designed for visual grounding, we derive bounding boxes from the provided segmentation masks or region annotations. All datasets are randomly split into training and testing subsets with an 8:2 ratio. Please refer to the appendix for more details.

\noindent \textbf{Implementation Details.} We implement MedLoc-R1 based on the Qwen2.5-VL family \cite{bai2024qwen}, with Qwen2.5-VL-3B-Instruct as our primary reasoning model, and additionally evaluate 7B and 32B variants to study scalability.
Following previous work~\cite{shen2025vlm,zheng2025easyr1}, we implement the code in Pytorch using 4 NVIDIA H800 80G GPUs. We follow the default GRPO setup during RL fine-tuning, with group size G set to 8, temperature to 0.9, and KL divergence ratio $\beta$ to 0.4. We use the AdamW optimizer with an initial learning rate of 1e-6 for both SFT and RL, a batch size of 1 per GPU, and 2-step gradient accumulation. 
The adaptive IoU threshold $\tau_k$ is initialized at \(\tau_0{=}0.3\) and gradually increased toward \(\tau_{\text{target}}{=}0.8\) when using the piecewise decay schedule with step sizes \(\delta^{(1)}{=}0.15\), \(\delta^{(2)}{=}0.10\), \(\delta^{(3)}{=}0.05\) and stage boundaries \(\beta^{(1)}{=}0.55\), \(\beta^{(2)}{=}0.75\), with additional results for alternative piecewise configurations reported in the Appendix. For the linear and cosine decay variants, we set \(\tau_0{=}0.2\), \(\tau_{\text{target}}{=}0.8\), and \(\delta_0{=}0.2\). A sliding window of length \(N{=}30\) is used to track performance statistics for all schedules.
All models are trained for up to 5 epochs, with evaluation conducted at step 1000.

\noindent \textbf{Evaluation Metrics.}
Our model outputs bounding boxes without confidence scores, making standard detection metrics like AP inapplicable. We instead use two metrics: Accuracy at specific IoU thresholds (A@0.5 and A@0.8), and a pseudo-mAP, computed as the mean accuracy over 10 evenly spaced thresholds from 0.5 to 0.95, i.e., $\text{mAP} = \frac{1}{K} \sum_{k=1}^{K} \text{Acc}_{\tau_k}$ with $K{=}10$. We run all experiments with three random seeds and report the average and standard deviation, and do significance analysis via paired two-tailed t-tests against each baseline under the same seeds.

\noindent \textbf{Baselines.}
We compare our method against five baselines.
(1) \textbf{Zero-shot}: directly evaluating the pretrained Qwen2.5-VL models without fine-tuning to assess their inherent grounding capability.
(2) \textbf{SFT}: supervised fine-tuning with LLaMA Factory~\cite{zheng2024llamafactory} to regress bounding boxes, without any RL or reward scheduling.
(3) \textbf{Raw-IoU}: GRPO trained with the continuous IoU score as the reward, removing any discrete thresholding.
(4) \textbf{Fixed-threshold (VLM-R1)}: following VLM-R1~\cite{shen2025vlm}, GRPO with a static IoU threshold $\tau_{\text{fixed}}\!=\!0.5$ to isolate the effect of dynamic thresholding.
(5) \textbf{V-Triune}: a re-implementation of the V-Triune-style schedule~\cite{ma2025vtriune}, which assigns three fixed thresholds to the early (0–10\%), middle (10–25\%), and late (25–100\%) training stages based only on progress. Each baseline is tuned to its best performance through extensive empirical validation within our setting for fairness.

\begin{figure*}[!t]
	\begin{center}
		\includegraphics[width=1\linewidth]{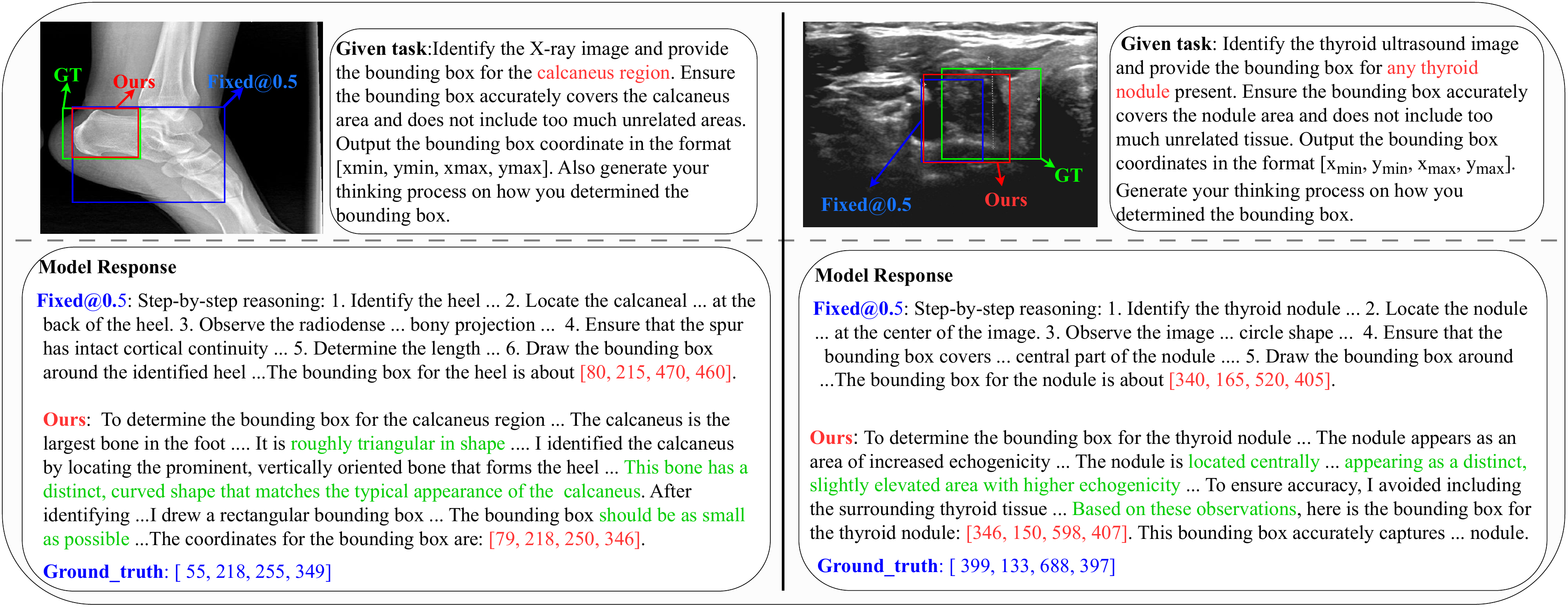}
	\end{center}
  	  \vspace{-0.4cm}
      \caption{Qualitative comparison of our MedLoc-R1 (in \textit{red} boxes) and fixed-threshold VLM-R1 (in \textit{blue} boxes) on HEEL and TN3K. Ground truth in \textit{green} boxes. MedLoc-R1 produces more precise boxes with coherent and semantically rich reasoning.}
    
    \label{fig:first} \vspace{-0.6cm}
\end{figure*}

\begin{figure}[t]
    \centering
    \includegraphics[width=1\linewidth]{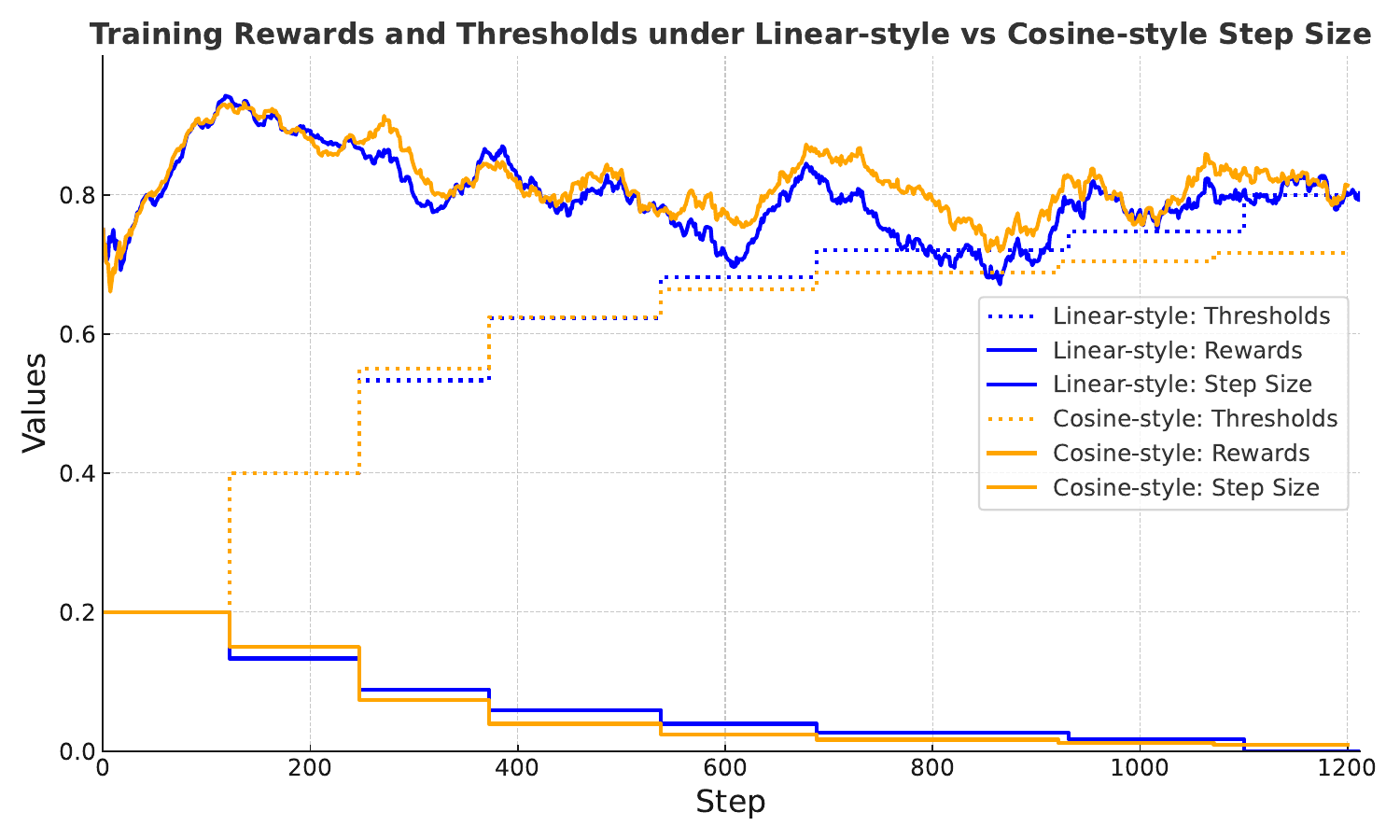}
    \vskip -0.2in
    \caption{\textbf{Visualization of how Linear- and Cosine-style step size} influence the evolution of thresholds and training rewards.}
    \label{fig:rewards_threshold_linear_consie_stpe} \vspace{-0.4cm}
    \vskip -0.2in
\end{figure}

\subsection{Experimental Results}

\noindent \textbf{Quantitative Results.}
Table~\ref{tab:main_table} reports the quantitative comparison across the three medical visual grounding datasets. MedLoc-R1 consistently outperforms all baselines under all evaluation metrics, with particularly large gains at the stricter A@0.8 threshold. It also scales effectively, providing consistent gains across 3B, 7B, and 32B models. Figure~\ref{fig:performance_over_stpes} further shows it converges faster and maintains more stable performance growth than the other baselines, supporting the effectiveness of our dynamic curriculum design. These confirm that performance-aware reward scheduling is crucial for stabilizing GRPO training in medical grounding.

\noindent \textbf{Comparison with baselines.}  
Zero-shot performance from pretrained vision-language models is notably weak across all metrics, highlighting the necessity of task-specific adaptation for accurate localization in medical domains. SFT substantially offers notable gains by directly optimizing bounding box predictions. However, it can not produce any reasoning process, directly predicting results. In clinical applications where transparency and trust are paramount, lacking interpretability is a critical shortcoming. Among RL-based baselines, Raw-IoU offers a continuous reward signal, but early-stage predictions yield minor score differences, providing insufficient contrast for effective GRPO updates. Fixed-threshold VLM-R1 is limited by the rigidity of its static reward boundary, yielding reward sparsity and vanishing gradients in early training and capping optimization potential in later stages. V-Triune takes a step further by updating thresholds according to a predefined schedule based solely on training progress, ignoring model performance and risking premature difficulty escalation and unstable optimization. MedLoc-R1 addresses these limitations by aligning reward difficulty with model readiness through performance-aware curriculum scheduling. Empirically, it achieves the strongest results (e.g., +43.18 A@0.8 on HEEL over VLM-R1 and +16.11 mAP on TN3K over V-Triune). Even when compared with Raw-IoU, MedLoc-R1 consistently delivers higher performance, illustrating the benefit of progressively tightening reward boundaries instead of relying on continuous signals with weak contrast. Additional baseline results can be found in the Appendix.

\noindent \textbf{Comparison of decay strategies.} 
We further compare the three step-size schedules in MedLoc-R1. Piecewise decay attains the highest performance but relies on multiple stage-specific step sizes and boundaries. In contrast, linear and cosine decay use only a single initial $\delta_0$ and still achieve competitive performance, suggesting that the simpler linear and cosine variants offer a favorable trade-off between performance and tuning complexity. Figure~\ref{fig:rewards_threshold_linear_consie_stpe} further illustrates these behaviors: linear decay tightens the threshold at a roughly constant pace, whereas cosine schedule closely mimics piecewise decay with larger threshold increments early and smaller ones later, producing slightly smoother reward and threshold trajectories. Despite these differences, all strategies support stable optimization and reach similar final performance, indicating that MedLoc-R1 can work well without extensive step-size tuning.

\noindent \textbf{Qualitative Results.}  Figure~\ref{fig:first} presents qualitative comparisons between MedLoc-R1 and the fixed-threshold baseline VLM-R1 ($\tau_{\text{fixed}}{=}0.5$) on HEEL and TN3K. In both examples, MedLoc-R1 produces bounding boxes that more closely match the anatomical target regions, while VLM-R1 often generates boxes that are either oversized or shifted away from clinically relevant structures. The reasoning traces further highlight the difference between the two methods. MedLoc-R1 identifies key visual cues such as the characteristic curvature of the calcaneus in HEEL and the echogenicity and positional patterns of the thyroid nodule in TN3K, and uses them to justify its localization. In contrast, VLM-R1 tends to provide lengthy procedural descriptions without integrating meaningful diagnostic evidence, which limits its spatial accuracy and interpretive usefulness. These qualitative observations align with the quantitative results, indicating that performance-aware reward scheduling helps the model focus on medically informative features during training. Representative failure cases are included in the appendix for completeness.

\begin{table}[t]
    \centering
    \small
    \caption{Ablation on HAM10000 showing the \textbf{effect of different update criteria}. \checkmark indicates the one is active for threshold update.}
    \vspace{-5pt}
    \setlength{\tabcolsep}{2pt}
    \begin{tabular}{lcccc}
    \toprule
    \textbf{Config} & \textbf{$\bar{r}_k \ge P_{\tau_k}$} & \textbf{$\bar{m}_k \ge \Delta$} & \textbf{$\sigma_{r,k} \le S_{\tau_k}$} & \textbf{A@0.5} \\
    \midrule
    Full (Ours)          & \cmark & \cmark & \cmark & \textbf{94.96} \\
    w/o Reward Check    & \xmark & \cmark & \cmark & 82.33 \\
    w/o IoU Margin Check     & \cmark & \xmark & \cmark & 90.86 \\
    w/o Stability Check & \cmark & \cmark & \xmark & 89.72 \\
    Only Reward Check   & \cmark & \xmark & \xmark & 88.32 \\
    Only IoU Margin Check     & \xmark & \cmark & \xmark & 84.92 \\
    Only Stability Check    & \xmark & \xmark & \cmark & 86.87 \\
    \bottomrule
    \end{tabular}
    \label{tab:ablation}
    \vskip -0.1in
\end{table}


\begin{table}[t]
\centering
\small
\captionof{table}{Ablation on \textbf{threshold scheduling strategies} on HAM10000. 
\textbf{Fixed-Aggressive} means it fixes a large $\delta_k$ and relaxed $P_k$/$S_k$, updating thresholds aggressively. The adaptive one dynamically tunes all three during training. \textbf{See Appendix for configuration details of \enquote{Dynamic}.}}
\vspace{-5pt}
\setlength{\tabcolsep}{4pt}
\begin{tabular}{lcccc}
\toprule
\textbf{Strategy} & \textbf{$\delta_k$} & \textbf{$P_k$} & \textbf{$S_k$} & \textbf{A@0.5} \\
\midrule
Adaptive (Ours)        & Dynamic & Dynamic & Dynamic & \textbf{94.96} \\
Fixed-Aggressive        & 0.15    & 0.60    & 0.40    & 71.54 \\
Fixed-Conservative      & 0.05    & 0.80    & 0.15    & 87.92 \\
Fixed-Moderate                & 0.10    & 0.70    & 0.25    & 83.92 \\
\bottomrule
\end{tabular}
\label{tab:param-strategy}
\vspace{-10pt}
\end{table}

\subsection{Ablation Study}

To validate the contributions of key components in our scheduling framework, we conduct ablation studies on HAM10000, focusing on three aspects: update criteria, scheduling strategy, and the sliding window mechanism.

\noindent \textbf{Effect of Update Criteria.} Table~\ref{tab:ablation} shows the effect of update criteria, including reward sufficiency, IoU margin adequacy, and reward stability. The full configuration, which requires all three to be satisfied before raising difficulty, achieves the best performance (94.96 A@0.5). Removing any individual component leads to large drops—most severely when omitting the reward check (–12.63), indicating its essential role in preventing premature updates under noisy signals. The IoU margin and stability checks contribute to filtering out uncertain or volatile learning phases. Notably, single-criterion variants fail to match the performance of dual or full configurations, confirming that all three criteria play complementary roles in triggering 
reliable, performance-aligned threshold progression.

\begin{figure}[t]
    \centering
    \begin{minipage}[t!]{0.48\linewidth}
        \centering
        \captionof{table}{\textbf{Ablation on Step Size Variants in Piecewise Decay} on HAM10000.\enquote{Identical} refers to a fixed step size $\delta_k = \delta_0$ throughout training.}
        \label{tab:delta_ablation}
        \vspace{-6pt}
        \resizebox{\linewidth}{!}{%
        \begin{tabular}{lc}
            \toprule
            Strategy & A@0.5 \\
            \midrule
            Dynamic-$\delta_k$ (Ours) & \textbf{94.96} \\
            Identical-$\delta_k=0.05$  & 76.29 \\
            Identical-$\delta_k=0.15$  & 79.64 \\
            Identical-$\delta_k=0.25$  & 77.13 \\
            \bottomrule
        \end{tabular}}
    \end{minipage}
    \hfill
    \begin{minipage}[t!]{0.48\linewidth}
        \centering
        \includegraphics[width=\linewidth]{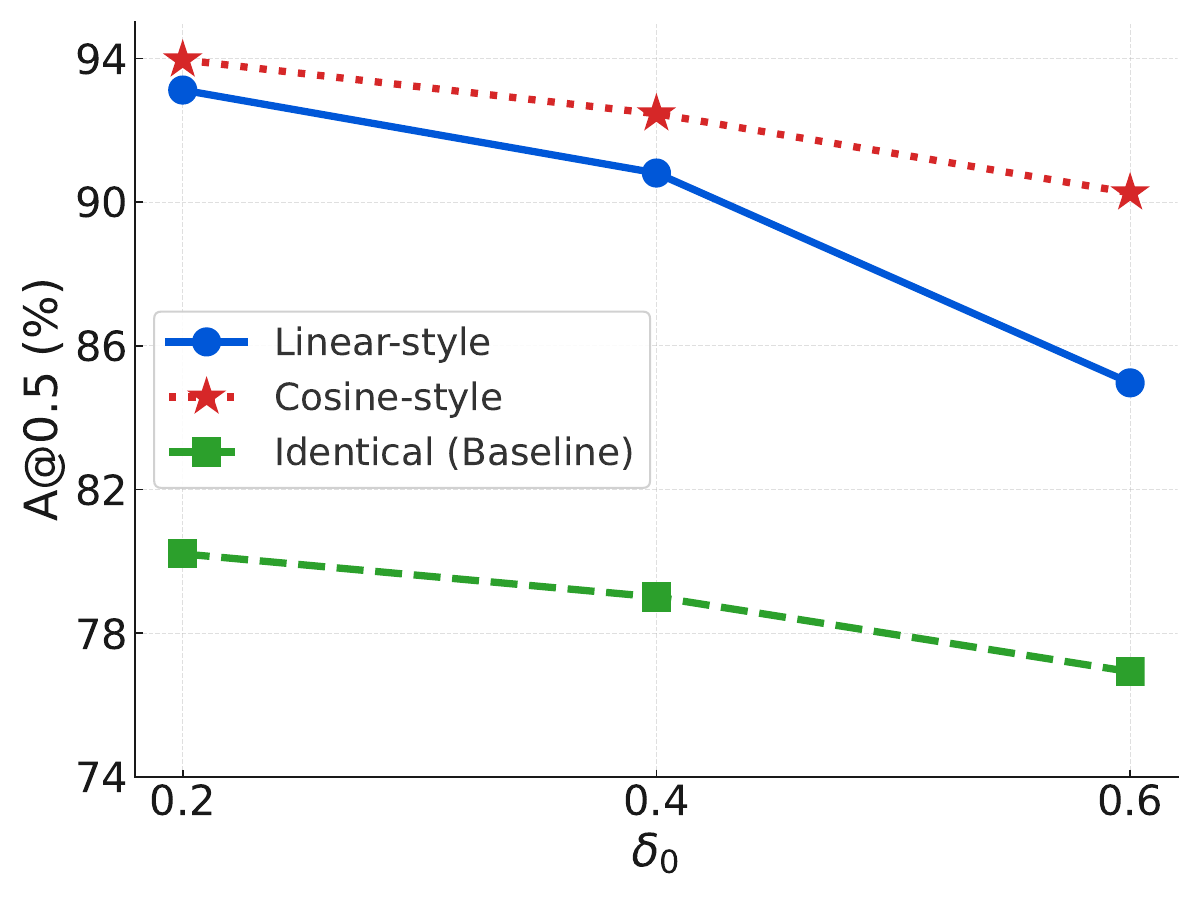}
        \vspace{-23pt}
        \captionof{figure}{Ablation on Step Size \textbf{$\delta_0$ of Linear vs. Cosine Decay} on HAM10000.}
        \label{fig:delta_ablation}
    \end{minipage}
\end{figure}

\noindent \textbf{Impact of Scheduling Strategy.}
We then ablate the scheduling strategy under the piecewise decay setting, comparing a dynamic schedule that adjusts $\delta_k$, $P_k$ and $S_k$ against fixed counterparts. As reported in Table~\ref{tab:param-strategy}, our method consistently outperforms all fixed counterparts, with a substantial margin of over 20\% compared to aggressive settings and 7--11\% over more conservative ones. 
We further isolates the effect of identical step size $\delta_k$ in Table~\ref{tab:delta_ablation}, revealing that static pacing, regardless of magnitude, leads to inferior results. 
Figure~\ref{fig:delta_ablation} then compares the initial step size $\delta_0$ for linear and cosine decay. Both variants outperform the identical baseline and cosine remains strongest.
Together, these findings suggest that MedLoc-R1 gains from Capability-aware scheduling, while supporting low-parameter decay schedules without extensive tuning.

\begin{figure}[t]
    \centering
    \includegraphics[width=0.9\linewidth]{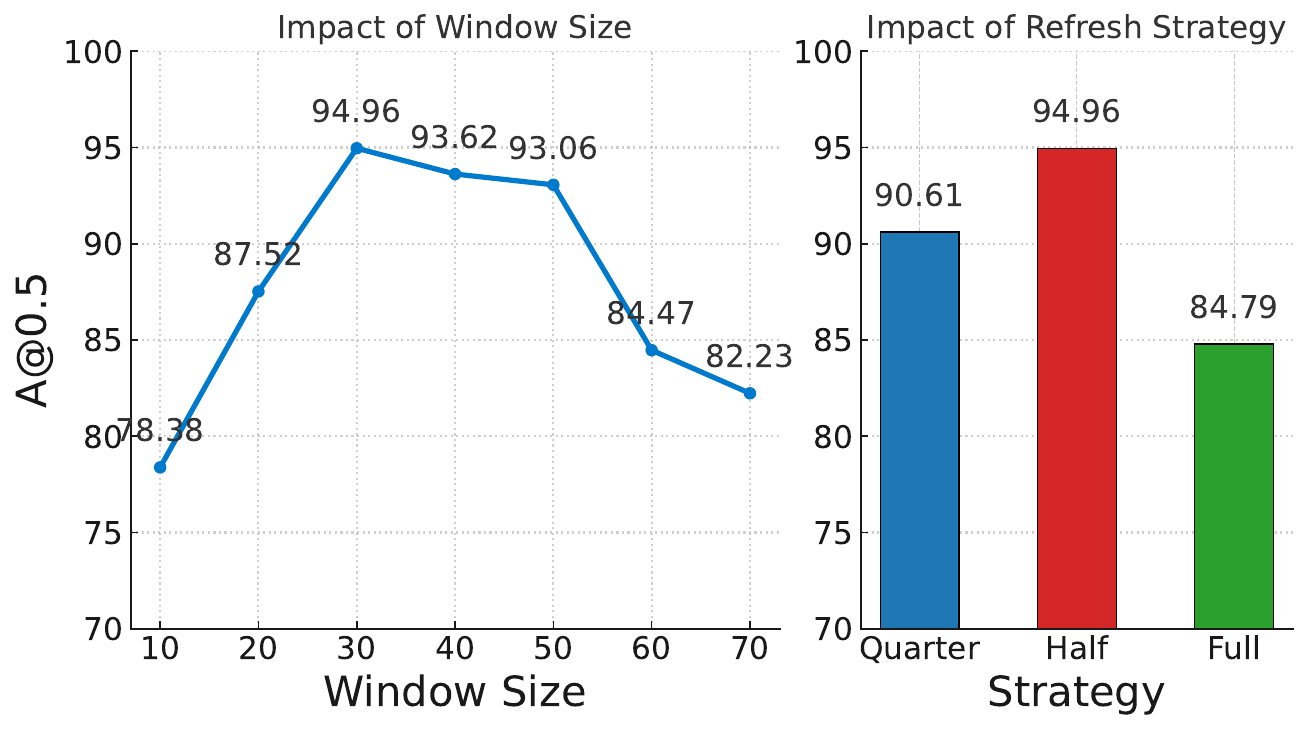}
    \vspace{-10pt}
    \caption{\textbf{Sliding Window Analysis:} Size \& Refresh Strategy. Left: A@0.5 across different window sizes. Right: effect of data refresh strategies once the threshold updates.}
    \label{fig:sliding-window-analysis} 
    \vskip -0.2in
\end{figure}

\noindent \textbf{Sliding Window Analysis.}
Lastly, we study the impact of the sliding window on
training trends. Figure~\ref{fig:sliding-window-analysis} shows that the window size significantly influences performance: small windows introduce noise and instability, while overly large ones hinder responsiveness. A size of 30 strikes the best balance between stability and reactivity. 
And the half-refresh strategy yields the highest accuracy, suggesting it best balances historical context and new information.

\section{Conclusion}
We proposed MedLoc-R1, a reinforcement learning framework for medical visual grounding with a progressive curriculum reward scheduling mechanism. By adaptively adjusting task difficulty according to model readiness, MedLoc-R1 alleviates reward sparsity, stabilizes training, and improves both grounding accuracy and explanation quality across medical imaging modalities. In future work, we will extend the framework to multi-task settings, such as joint lesion detection and disease classification.



{
    \small
    \bibliographystyle{ieeenat_fullname}
    \bibliography{main}
}

\clearpage
\setcounter{page}{1}
\maketitlesupplementary


\appendix
\section{Datasets Details}
We provide detailed statistics of the dataset splits for the three benchmarks used in our experiments: \textbf{HAM10000}, \textbf{HEEL}, and \textbf{TN3K}.

\begin{itemize}
  \item \textbf{HAM10000} is a public dataset containing 10,015 dermoscopic images released in 2018 by the Medical University of Vienna. It is one of the most important benchmark datasets in the field of automatic skin cancer detection and classification. It covers 7 common skin lesion types: actinic keratoses (akiec), basal cell carcinoma (bcc), benign keratosis-like lesions (bkl), dermatofibroma (df), melanoma (mel), melanocytic nevi (nv), and vascular lesions (vasc). We derived bounding boxes from the provided lesion masks and randomly split them into training and test sets with 8,012 training and 2,003 testing samples across 7 skin lesion categories.
  \item \textbf{HEEL} is a public dataset of 3{,}956 lateral foot X-ray images collected at Kirkuk General Hospital, comprising three diagnostic categories: \texttt{Normal} (1{,}842 images), \texttt{Heel Spur} (1{,}316 images), and \texttt{Sever's disease} (798 images). All images are labeled by orthopedic specialists and cross-validated by radiologists. Following the original protocol, we use 3{,}164 images for training and 792 for testing, while preserving the class distribution across the three categories.

  \item \textbf{TN3K} is an open-access thyroid nodule ultrasound dataset comprising 3{,}493 B-mode images from 2{,}421 patients, each annotated with pixel-wise nodule masks. The official split contains 2{,}879 training and 614 test images. In our experiments, we derive a binary classification subset with 2{,}655 training and 544 testing samples, labeled as \texttt{malignant} and \texttt{benign} cases.

\end{itemize}

To offer a clearer overview of their composition, Table~\ref{tab:dataset_stats} summarizes the class-wise training and testing splits for all three datasets.

\begin{table}[t]
  \centering
  \small
  \caption{Class-wise number of training and test samples for the three benchmarks.}
  \setlength{\tabcolsep}{6pt}
  \begin{tabular}{llrr}
    \toprule
    Dataset & Class      & Train & Test \\
    \midrule
    \multirow{8}{*}{HAM10000}
      & nv             & 5367 & 1338 \\
      & mel            &  900 &  213 \\
      & bkl            &  868 &  231 \\
      & bcc            &  418 &   96 \\
      & akiec          &  253 &   74 \\
      & vasc           &  118 &   24 \\
      & df             &   88 &   27 \\
      & \textbf{Total} & 8012 & 2003 \\
    \midrule
    \multirow{4}{*}{HEEL}
      & Normal         & 1473 &  369 \\
      & Heel Spur      & 1053 &  263 \\
      & Severe         &  638 &  160 \\
      & \textbf{Total} & 3164 &  792 \\
    \midrule
    \multirow{3}{*}{TN3K}
      & malignant      &  932 &  219 \\
      & benign         & 1723 &  331 \\
      & \textbf{Total} & 2655 &  544 \\
    \bottomrule
  \end{tabular}
  \label{tab:dataset_stats}
\end{table}

\section{Experimental Configuration Details}

We elaborate here on the configuration details of the threshold scheduling strategies evaluated in Table~3. All strategies manipulate three key scheduling parameters: the step size $\delta_k$, the performance percentile $P_k$, and the stability margin $S_k$. These parameters govern how the threshold $\tau_k$ is updated during training.

\paragraph{Adaptive (Ours).}  
The adaptive strategy \textbf{dynamically} adjusts $\delta_k$, $P_k$, and $S_k$ according to the current threshold value $\tau_k$ throughout training. It operates in three regimes:
\begin{itemize}
    \item When $\tau_k < 0.60$, a large step size $\delta_k = 0.15$ is used to encourage rapid threshold progression, with $P_k = 0.80$ and $S_k = 0.20$.
    \item When $0.60 \leq \tau_k < 0.75$, we moderate the update using $\delta_k = 0.10$, $P_k = 0.75$, and $S_k = 0.35$.
    \item When $\tau_k \geq 0.80$, updates become conservative with $\delta_k = 0.05$, $P_k = 0.55$, and $S_k = 0.40$.
\end{itemize}
This staged configuration allows the model to explore aggressively in early training while stabilizing and refining predictions in later stages, leading to more robust convergence behavior.

\paragraph{Fixed-Aggressive.}  
This strategy uses fixed values across the entire training process, with a large step size $\delta_k = 0.15$, a relatively low performance percentile $P_k = 0.60$, and a small stability margin $S_k = 0.40$. The configuration encourages fast threshold updates with minimal stability constraints, favoring aggressive adaptation dynamics.

\paragraph{Fixed-Moderate.}  
The moderate variant sets intermediate values: $\delta_k = 0.10$, $P_k = 0.70$, and $S_k = 0.25$. It aims to balance adaptation speed and stability, representing a middle ground between aggressive and conservative strategies.

\paragraph{Fixed-Conservative.}  
This configuration employs a small step size $\delta_k = 0.05$, high performance requirement $P_k = 0.80$, and a tight stability margin $S_k = 0.15$. These constraints slow down the threshold adaptation process, ensuring greater caution and smoother updates throughout training.

All fixed strategies retain constant values for all three parameters, while the adaptive strategy transitions between configurations in a stage-wise manner depending on $\tau_k$. This dynamic scheduling is a key factor contributing to the superior performance of our method.

\section{More results on Piecewise Decay Schedule}
To assess how the hyperparameters of the piecewise decay schedule influence RL-based localization, we perform an ablation over the stage-wise step sizes $(\delta^{(1)},\delta^{(2)},\delta^{(3)})$ and the boundary thresholds $(\beta^{(1)},\beta^{(2)})$. 
In all settings, the adaptive IoU threshold is initialized at $\tau_0{=}0.3$ and driven toward $\tau_{\text{target}}{=}0.8$, while the schedule parameters control \emph{how fast and in what shape} this transition occurs. 
Our main results adopt the configuration $\delta^{(1)}{=}0.15$, $\delta^{(2)}{=}0.10$, $\delta^{(3)}{=}0.05$ and $\beta^{(1)}{=}0.55$, $\beta^{(2)}{=}0.75$; 
Table~\ref{tab:abl_piecewise} reports additional piecewise configurations evaluated on HAM10000, HEEL, and TN3K. 
The default choice consistently achieves the best or near-best A@0.5, whereas both more aggressive and flatter step-size patterns lead to noticeable drops in performance. 
These results indicate that the schedule is not overly sensitive within a reasonable range, but benefits from \emph{moderate early-stage increments followed by gentler late-stage refinement}, which provides a stable progression of $\tau_k$ and alleviates reward sparsity during training.

\begin{table}[t]
\centering
\small
\caption{Ablation of \textbf{piecewise decay} schedule parameters.}
\setlength{\tabcolsep}{6pt}
\begin{tabular}{lccccr}
\toprule
Dataset & $\delta_1$ & $\delta_2$ & $\beta_1$ & $\beta_2$ & A@0.5 \\
\midrule
\multirow{4}{*}{HAM (0.3--0.8)}
& 0.10 & 0.05 & 0.55 & 0.70 & 92.23 \\
& 0.15 & 0.10 & 0.55 & 0.75 & \textbf{94.26} \\
& 0.20 & 0.15 & 0.60 & 0.75 & \underline{93.97} \\
& 0.25 & 0.20 & 0.60 & 0.75 & 93.10 \\
\midrule
\multirow{4}{*}{HEEL (0.3--0.8)}
& 0.10 & 0.05 & 0.50 & 0.70 & 93.38 \\
& 0.15 & 0.10 & 0.50 & 0.75 & \textbf{94.19} \\
& 0.20 & 0.15 & 0.60 & 0.75 & \underline{94.11} \\
& 0.25 & 0.20 & 0.60 & 0.75 & 93.87 \\
\midrule
\multirow{4}{*}{TN3K (0.3--0.8)}
& 0.10 & 0.05 & 0.50 & 0.70 & 65.97 \\
& 0.15 & 0.10 & 0.50 & 0.75 & \underline{66.18} \\
& 0.20 & 0.15 & 0.60 & 0.75 & \textbf{66.31} \\
& 0.25 & 0.20 & 0.60 & 0.75 & 65.56 \\
\bottomrule
\end{tabular}
\label{tab:abl_piecewise}
\end{table}

\section{Additional baseline results}
To provide a stronger empirical comparison, we additionally included three external baselines from recent literature. \textbf{GroundingDINO-L}~\cite{gdino} is a state-of-the-art open-set object detector, which we fine-tuned on our training set and used the highest-confidence predicted box as the final prediction. \textbf{BoxMed-RL}~\cite{jing2025BoxMed-RL}, originally developed for radiology report generation, adopts GRPO-based Spatially Verifiable Reinforcement (SVR) to align medical findings with bounding boxes on sentence-box aligned datasets. Its IoU-based reward, defined as IoU when IoU \(>0\) and 0 otherwise, is effectively equivalent to our Raw-IoU baseline. We therefore reproduced its SVR framework in our bbox-only setting using instruction prompts of the form "Provide the bounding box for \{target\}." \textbf{MedGround-R1}~\cite{xu2025medground} is a recent GRPO-based medical grounding method that combines spatial accuracy and semantic consistency in the reward design, together with a Chain-of-Box reasoning template. As shown in Table~\ref{tab:main_and_ablation} (left), MedLoc-R1 consistently outperforms all three external baselines across the three datasets in terms of A@0.5, further validating the effectiveness of our performance-aware curriculum reward scheduling. In addition, Table~\ref{tab:main_and_ablation} (right) reports a grid search over the key GRPO hyperparameter, namely the group size \(G\).

\begin{table}[t]
  \centering
  \caption{Additional baseline results (\textbf{left}) and main ablation study on group size G (\textbf{right}).}
  \label{tab:main_and_ablation}
  \small
  \scriptsize
  \setlength{\tabcolsep}{4pt}

  \begin{subtable}[t]{0.68\columnwidth}
    \centering
    
    \label{tab:main_results}
    \begin{tabular}{lccc}
      \toprule
      Method & HAM10000 & HEEL & TN3K \\
      \midrule
      GroundingDINO-L & 84.27 & 83.61 & 32.70 \\
      BoxMed-RL       & \underline{92.36} & 74.51 & \underline{56.39} \\
      MedGround-R1    & 88.23 & \underline{89.52} & 51.43 \\
      MedLoc-R1 (Ours)       & \textbf{94.46} & \textbf{94.19} & \textbf{66.18} \\
      \bottomrule
    \end{tabular}
  \end{subtable}
  \hfill
  \begin{subtable}[t]{0.30\columnwidth}
    \centering
    \label{tab:ablation_groupsize}
    \begin{tabular}{cc}
      \toprule
      Group size & HAM10k \\
      \midrule
      4  & 89.10 \\
      6  & 93.51 \\
      8  & \textbf{94.46} \\
      10 & 94.36 \\
      \bottomrule
    \end{tabular}
  \end{subtable}
  \vspace{-10pt}
\end{table}


\begin{table}[t]
\centering
\footnotesize
\caption{Performance comparison on medical imaging datasets}
\label{tab:results}
\resizebox{\columnwidth}{!}{
\begin{tabular}{l|ccc|ccc}
\toprule
\multirow{3}{*}{Method} & \multicolumn{3}{c|}{HAM10000} & \multicolumn{3}{c}{HEEL}\\
& A@0.5 & A@0.8 & mAP & A@0.5 & A@0.8 & mAP \\
\midrule
Qwen-2.5-VL-3B & 35.75 & 3.39 & 12.38 & 13.64 & 1.64 & 4.78 \\
InternVL-2.5-4B & 33.19 & 2.31 & 11.23 & 11.17 & 1.01 & 3.81 \\
MedLoc-3B (Qwen) & 93.96 & 72.79 & 70.71 & 93.31 & 47.67 & 57.42 \\
MedLoc-4B (InternVL) & 90.03 & 69.85 & 67.80 & 90.97 & 45.19 & 55.46 \\
\bottomrule
\end{tabular}
}
\end{table}

\begin{figure*}[t]
	\begin{center}
		\includegraphics[width=0.9\linewidth]{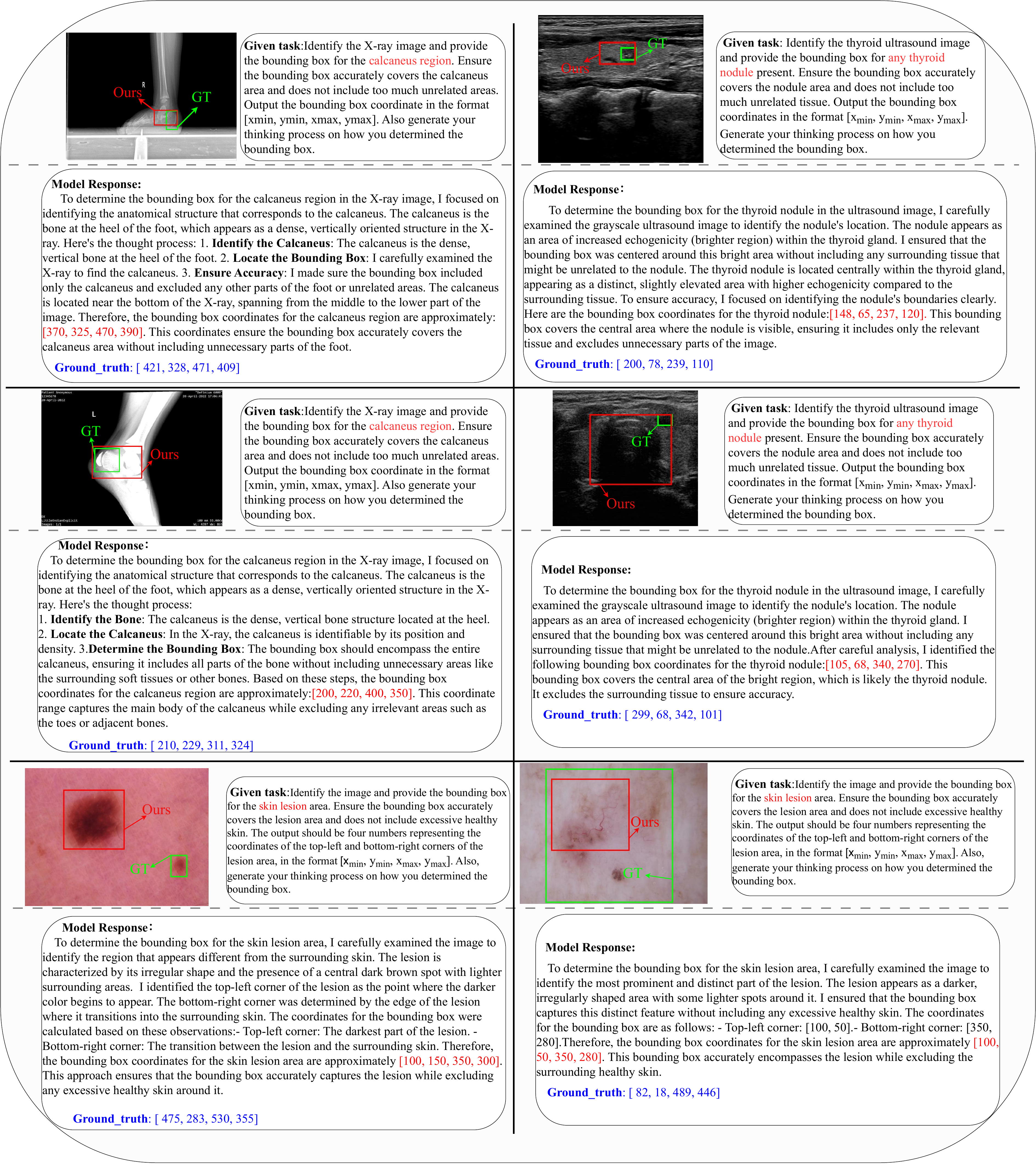}
	\end{center}
  	  \vspace{-0.4cm}
      \caption{
    Visualization of representative failure cases from our method on the HEEL, TN3K, and HAM10000 datasets. Red boxes denote the predicted bounding boxes by our MedLoc-R1 method, while green boxes indicate the ground truth. 
}

    \label{fig:failure} \vspace{-0.6cm}
\end{figure*}

\section{Failure Case Visualizations}
We present several representative failure cases of MedLoc-R1 in Figure~\ref{fig:failure}. Although the predicted bounding boxes are not always perfectly aligned with the ground truth, they are generally centered on the correct target regions, indicating that the model captures the key visual and semantic cues required for medical grounding.

On HEEL, MedLoc-R1 localizes the calcaneus region with good anatomical consistency. In failure cases, the predicted boxes may slightly under-cover or over-cover the annotated region, but they still focus on the correct bone structure while excluding most irrelevant areas. On TN3K, some predictions extend beyond the nodule boundary or miss subtle margins, yet the boxes remain centered on the relevant thyroid nodule region, suggesting that the model effectively exploits both spatial and intensity cues despite the ambiguity of ultrasound images. On HAM10000, although the predicted boxes may omit faint peripheral areas or include limited surrounding healthy skin, they generally cover the diagnostically important part of the lesion.

Overall, these examples suggest that the primary failure mode of MedLoc-R1 lies in spatial imprecision rather than incorrect target identification. Even when the localization is not exact, the model usually attends to the appropriate anatomical or pathological region, further supporting its effectiveness in medical visual grounding.

\end{document}